\documentclass[letterpaper, 10 pt, conference]{ieeeconf}  

\IEEEoverridecommandlockouts                              

\usepackage{graphics} 
\usepackage{amsmath} 
\usepackage{amssymb}  
\usepackage{algorithm}
\usepackage{algorithmic}
\usepackage{amsmath}
\usepackage{setspace}
\usepackage{tabularx}
\usepackage{booktabs} 
\usepackage{graphicx}

\usepackage{subcaption}
\usepackage[table]{xcolor}
\usepackage[colorlinks = true,
            linkcolor = black,
            urlcolor  = blue,
            citecolor = magenta,
            bookmarks = true]{hyperref}
\usepackage{comment}
\usepackage{todonotes}
\usepackage{multirow}
\usepackage{pgfplots}
\usepackage{pgfplotstable}
\pgfplotsset{compat=1.9}

\newcommand{\figurename}{Fig.}

\DeclareMathOperator{\sort}{sort}
\DeclareMathOperator{\OVRS}{OVRS}
\DeclareMathOperator{\countop}{count}

\title{\LARGE \bf
EEG-Driven Intention Decoding: Offline Deep Learning Benchmarking on a Robotic Rover
}

\author{
Ghadah Alosaimi\textsuperscript{1,2},
Maha Alsayyari\textsuperscript{2,4},
Yixin Sun\textsuperscript{2},
Stamos Katsigiannis\textsuperscript{2},\\
Amir Atapour-Abarghouei\textsuperscript{2},
Toby P. Breckon\textsuperscript{2,3} \\[0.5em]
\textsuperscript{1}Department of Computer Science, Imam Mohammad Ibn Saud Islamic University, Saudi Arabia \\
Department of \{\textsuperscript{2}Computer Science, \textsuperscript{3}Engineering\}, Durham University, UK \\
\textsuperscript{4}Department of Computer Science, King Saud University, Saudi Arabia
}

\begin{document}
\maketitle
\thispagestyle{empty}
\pagestyle{empty}

\begin{abstract}

Brain–computer interfaces (BCIs) provide a hands-free control modality for mobile robotics, yet decoding user intent during real-world navigation remains challenging. This work presents a brain–robot control framework for offline decoding of driving commands during robotic rover operation. A 4WD Rover Pro platform was remotely operated by 12 participants who navigated a predefined route using a joystick, executing the following commands: forward, reverse, left, right, and stop. Electroencephalogram (EEG) signals were recorded with a 16-channel OpenBCI cap and aligned with motor actions at $\Delta = 0$ ms and eight future prediction horizons ($\Delta > 0$ ms). After data preprocessing, eleven deep learning (DL) models were benchmarked for the task of intent classification, across the Convolutional Neural Network (CNN), Recurrent Neural Network (RNN), and Transformer  architectural families. ShallowConvNet achieved the highest performance for both action prediction (F1-score 67\% at $\Delta = 0$ ms) and intent prediction  (F1-score 66\% at $\Delta = 300$ ms), maintaining robust performance at future horizons. 
By combining real-world robotic control with multi-horizon EEG intention decoding, this study introduces a reproducible benchmark and reveals key design insights for predictive, DL-based BCI systems.

\end{abstract}
\section{INTRODUCTION} \label{s:introduction}

Brain–computer interfaces (BCIs) enable direct communication between the human brain and external systems, with applications spanning assistive mobility, intelligent vehicles, and human–robot interaction~\cite{maiseli2023brain}, among others. Among these, brain-controlled vehicles (BCVs) have gained significant traction as platforms where neural signals can supplement or even replace manual control for navigation and decision-making. BCVs are generally classified into two categories: servo-level systems, which translate brain activity into low-level actuator commands (e.g., throttle or steering), and task-level systems, which decode higher-order driving intentions such as braking, turning, or path selection~\cite{lu2018eeg}. Human adaptability and contextual awareness are indispensable in dynamic traffic scenarios, underscoring the importance of intention-aware BCIs in BCVs. By decoding driver intentions, such systems can enable shared autonomy and neuroadaptive control that combine human flexibility with machine precision~\cite{9499083}.

\begin{figure*}[!]
    \centering
    \includegraphics[width=0.90\textwidth]{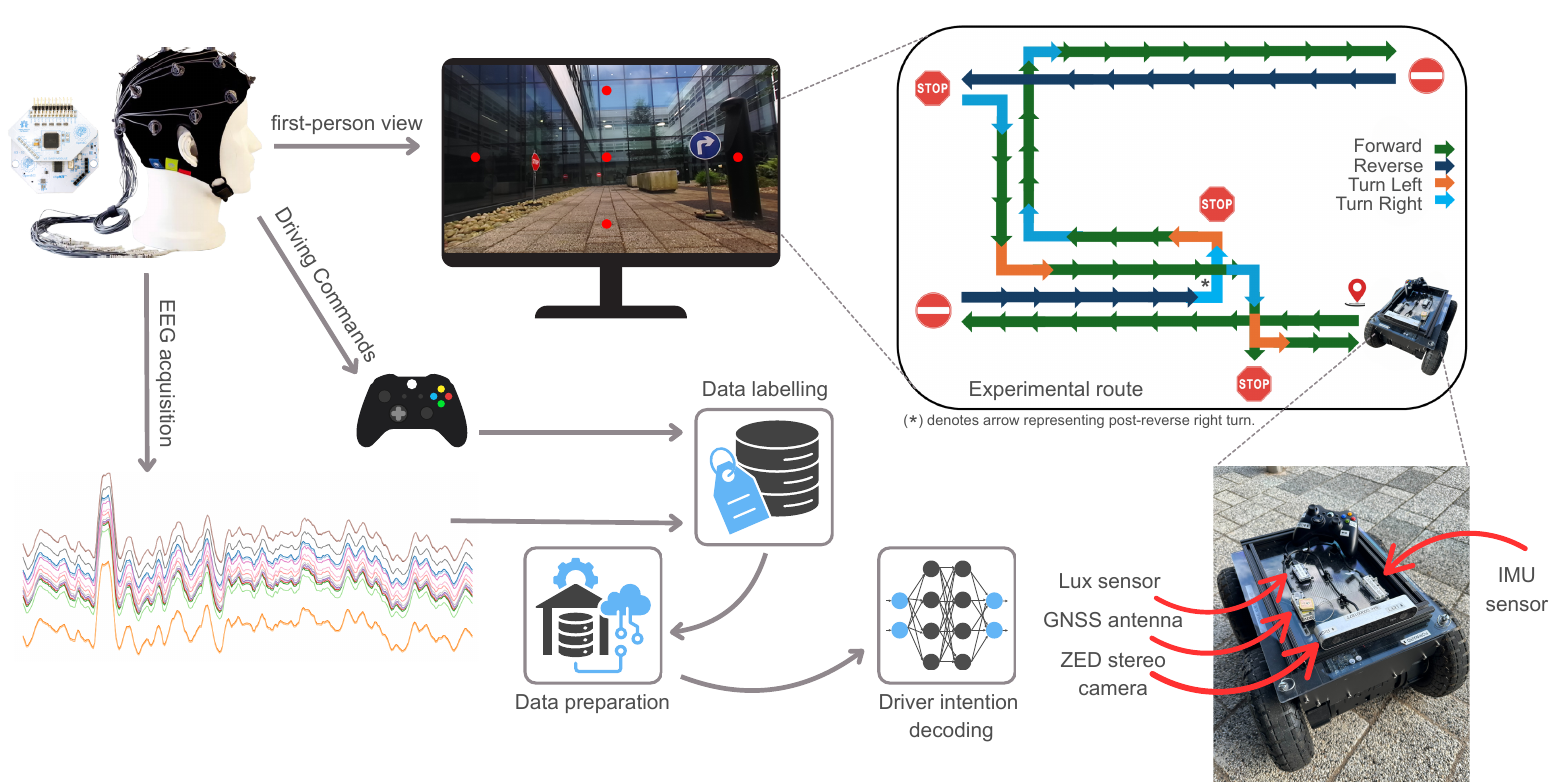}
    \caption{Experimental setup for EEG-based driver intention prediction. 
    EEG signals were acquired from the subject and synchronised with a first-person 
    view of the route. Commands were sent 
    via a controller and recorded for data labelling. The rover was equipped with a ZED stereo camera, GNSS antenna, Lux, and IMU sensors to capture 
    multimodal information during the experimental route. The recorded data were processed through preparation and labelling pipelines before being used for driver intention decoding.} 
    \label{fig:exp_setup}
\end{figure*}

Although several electroencephalogram (EEG) paradigms have been applied in BCI-based driving, including motor imagery (MI), P300, and steady-state visual evoked potentials (SSVEP)~\cite{cruz2021self, lu2019model}, most prior work remains limited to controlled laboratory settings, often relying on virtual simulators and prototype platforms~\cite{mezzina2020four, 10055923}. Furthermore, prior studies have primarily focused on discrete commands or single maneuver types (e.g., braking), with limited exploration of multi-command decoding or predictive intention recognition across future time horizons~\cite{yao2024enhancing, lutes2025few}. While deep learning  (DL) has significantly advanced EEG decoding performance~\cite{altaheri2023deep}, few studies have systematically compared diverse DL architectures under consistent preprocessing and evaluation protocols, and transformer-based models remain largely unexplored in BCI-driven driving contexts~\cite{vafaei2025transformers}.

To address these limitations, this study introduces a real-world brain–robot control framework for offline intention decoding. In our setup (Fig.~\ref{fig:exp_setup}), participants operated a mobile robotic rover along a predefined outdoor path, while executing five navigational commands: forward, reverse, left, right, and stop. Neural activity was recorded using a 16-channel gel-free EEG headset~\cite{openbci} and temporally aligned with motor labels derived from joystick input for real-time decoding ($\Delta$ = 0 ms) and anticipatory decoding at future time horizons ($\Delta \in [300, 1000]$ ms). EEG recordings were preprocessed through a standardised pipeline and segmented into overlapping windows for model input. A comprehensive DL benchmark was conducted across 11 architectures, including convolutional, recurrent, and transformer-based models. All evaluations employed temporally stratified splits and training-only oversampling to ensure label integrity and minimise data leakage to facilitate robust multi-command classification under realistic conditions. Results show that compact CNNs, especially ShallowConvNet, consistently outperformed other architectures, achieving a robust F1-score of 66\% at the $\Delta = 300$ ms prediction horizon. \smallskip

\noindent
The main contributions of this work are as follows:

\begin{itemize}
\item A realistic BCV experiment is designed and implemented, combining outdoor rover remote navigation with synchronised EEG–behaviour labelling of five driving commands (forward, reverse, left, right, and stop) under real-world conditions. 
\item A temporal-aware label-stratified evaluation pipeline is introduced to mitigate temporal leakage and support robust offline benchmarking of action and intention prediction.
\item A systematic comparison of 11 state-of-the-art DL models is conducted, including transformer-based models, establishing rigorous performance baselines for multi-session, multi-command EEG-based BCV decoding.
\end{itemize}

\section{Related Work} \label{related_works}
A number of studies have investigated BCIs for mobile robot control using various EEG paradigms and machine learning (ML) techniques. This section reviews relevant literature across three key themes: intention decoding in driving tasks (Section \ref{sec:lit_rev:intention}), the contrast between real-world and simulated experimental setups (Section \ref{sec:lit_rev:drive_scenarios}), and the use of DL for EEG decoding (Section \ref{sec:lit_rev:eeg_dl}). These thematic areas help situate the contributions of the present study within the broader context of BCI-driven navigation systems.

\subsection{Intention Decoding in Driving Scenarios}
\label{sec:lit_rev:intention}
Intention decoding from EEG has been a central topic in BCI research for BCVs, especially for interpreting directional or braking-related commands. Mezzina et al.~\cite{mezzina2020four} used a P300-based spatio-temporal interface to enable four-wheel prototype robotic car control, demonstrating promising accuracy for discrete commands in a structured indoor setting. Hekmatmanesh et al.~\cite{hekmatmanesh2021imaginary} demonstrated the feasibility of MI-based control for mobile vehicles using DL but their experiments were conducted in a controlled laboratory environment with restricted movement tasks and artificial visual feedback.  Several studies have attempted to extend BCI control to multiple navigation intentions, though typically within constrained setups. Gougeh et al.~\cite{gougeh2019automatic} proposed a MI-based framework to classify (left, right, and brake) intentions, achieving high offline accuracy (94.6\%) using spatial filtering and Support Vector Machines (SVM), but limited to symbolic cues and simulated tasks without real driving integration. Gui et al.~\cite{gui2024brain} developed an SSVEP-based brain-controlled mobile robot system combining task-related component analysis, deep neural networks, and model predictive control, to enable three navigation commands (left, right, and forward) on a TurtleBot4 with real-time validation in both simulated and real environments. Liu et al.~\cite{liu2021enhanced} introduced a generative adversarial network (GAN)-based augmentation framework to improve MI-based teleoperation (left and right) of a construction Unmanned Ground Vehicle, reporting gains of 4\% in accuracy, yet evaluations remained offline using a very small sample size. Recent work has also shifted towards anticipatory decoding. For example, Lutes et al.~\cite{lutes2025few} employed convolutional spiking neural networks (CSNN) to detect braking intention up to one second in advance, offering a predictive framework based on pre-movement brain potentials. However, the majority of prior studies remain constrained to binary or three-class setups, with few attempts at decoding richer command sets. Most studies focus only on short-term prediction of a manoeuvre, while continuous multi-command intention decoding under real-world driving conditions remains largely unaddressed~\cite{di2021robust, nguyen2019detection}. Motivated by these gaps, this study evaluates continuous multi-command intention decoding from EEG under realistic robotic navigation conditions.

\subsection{Real-world vs. Simulated Driving Scenarios}
\label{sec:lit_rev:drive_scenarios}
Beyond the type of commands decoded, another key limitation lies in the real-world relevance of experimental setups. Most BCV studies are still performed in virtual or indoor environments. Hoshino et al.~\cite{hoshino2021brain} used CNNs to classify control commands from EEG in a lab-based setup, relying on visual attention tasks. Lian et al.~\cite{lian2023novel} developed a novel asynchronous driver–vehicle interface in a CarSim simulator for left–right navigational control through EEG, with continuous two-class decoding accuracy surpassing 83\%. Lutes et al.~\cite{lutes2025few} further evaluated anticipatory braking with CSNNs in a simulator, while Zou et al.~\cite{zou2023mi} tested hybrid paradigms indoors with controlled lighting and displays. In contrast, only a few studies, such as Lucas et al.~\cite{lucas2022first}, have implemented BCIs for robotic vehicle control, where participants navigated obstacle courses using SSVEP-based commands, though no quantitative accuracy metrics were reported. Chang et al.~\cite{chang2022driving} investigated steering processes using multilayer dynamic brain network analysis during driving, providing insight into EEG connectivity but not into practical decoding accuracy. Although these studies demonstrate the feasibility of mobile BCIs, most do not combine real-world navigation, multi-command decoding, and predictive labelling under realistic conditions. In contrast, this study evaluates EEG-based decoding during outdoor robotic rover operation, providing a reproducible benchmark in real-world conditions.

\subsection{Deep Learning for EEG Decoding}
\label{sec:lit_rev:eeg_dl}
Earlier BCI studies have predominantly relied on traditional ML methods such as Common Spatial Pattern, Linear Discriminant Analysis, and SVM~\cite{aljalal2020comprehensive, rashid2020recent}. While effective in some cases, these approaches struggle with the high dimensionality and non-stationarity of EEG signals and require extensive handcrafted feature engineering. DL offers a solution by jointly learning spatial and temporal representations directly from raw or minimally processed EEG. Systematic reviews by Altaheri et al.~\cite{altaheri2023deep} and Hossain et al.~\cite{hossain2023status} highlight the success of CNNs, RNNs, and Transformers in MI and SSVEP decoding, though most evaluations remain offline or simulator-based. Moreover, few studies have compared diverse DL architectures under consistent preprocessing and evaluation pipelines, leaving open questions about their robustness in real-world multi-command navigation~\cite{9499083}. To address this gap, this study systematically benchmarks 11 state-of-the-art models, including CNNs (EEGNet~\cite{lawhern2018eegnet}, ShallowConvNet~\cite{schirrmeister2017deep}, DeepConvNet~\cite{schirrmeister2017deep}, CCNN~\cite{yang2018continuous}, CNN1D~\cite{taghizadeh2024eeg}, TSCeption~\cite{ding2022tsception}, STNet~\cite{zhang2022ganser}), RNNs (LSTM~\cite{zhang2021deep}, GRU~\cite{zhang2021deep}), and transformers (EEG-Conformer~\cite{song2022eeg}, ViT~\cite{dosovitskiy2020image}) in a real-world driving intention decoding context.
\section{Materials and Methods}
This section outlines the experimental design, including details of participant recruitment (Section \ref{sec:method:participants}), the robotic navigation framework (Section \ref{sec:method:exp_setup}), data labelling strategy (Section \ref{sec:method:labelling}), EEG preprocessing pipeline (Section \ref{sec:method:preparation}), model training and benchmarking (Section \ref{sec:method:model_train}), and evaluation metrics (Section \ref{sec:method:metrics}).

\subsection{Participants}
\label{sec:method:participants}
Twelve healthy adults (6 males, 6 females; aged 20–40) participated in the study by sitting in front of a screen and remotely operating a mobile robotic rover along a predefined outdoor path, while EEG was recorded. All participants had normal or corrected-to-normal vision and no history of neurological or psychiatric disorders. The study was approved by the Durham University Institutional Review Board under approval reference COMP-2023-04-03T15\_17\_29. All participants provided written informed consent prior to data collection. Each participant completed 10 sessions across multiple days, with each session lasting approximately 20 minutes, yielding a multi-session dataset for offline analysis.

\subsection{Experimental Setup}
\label{sec:method:exp_setup}

The experimental framework was originally designed as an indoor visual stimulation and EEG acquisition system synchronised with an outdoor robotic navigation task. Data collection followed an SSVEP-style setup, where participants were seated 90 cm from a primary display showing a real-time first-person view (FPV) from the rover. Five flickering red circular stimuli (\figurename~\ref{fig:exp_setup}) were overlaid in a cross (“+”) layout, each mapped to a driving command: stop (5 Hz, top circle), left (10 Hz, left circle), reverse (15 Hz, middle circle), right (20 Hz, right circle), and forward (30 Hz, bottom circle). Stimuli were rendered in PsychoPy ~\cite{peirce2019psychopy2} on a 60 Hz LCD screen with equal luminance, spacing, and size, and the colour red was chosen given evidence that long-wavelength warm hues enhance SSVEP amplitudes and improve classification ~\cite{chu2016influence}. A secondary monitor
allowed the supervising researcher to monitor impedance
levels and EEG quality in real time. Although these ancillary SSVEP signals were recorded, they were not analysed in this study. Instead, continuous EEG collected during outdoor rover driving was used directly for multi-command intention decoding. Participants navigated the robot along a predetermined route using an Xbox controller. The experimental route was manually designed to include straight segments, left/right turns, stops, and reverse manoeuvres. The same fixed route was used for all participants to ensure consistency across sessions. The platform was a 4WD Rover Pro (Rover Robotics), equipped with the following sensors mounted on a custom payload to enable reliable multimodal data acquisition in outdoor environments (see \figurename~\ref{fig:exp_setup}):

\begin{itemize}
\item \textbf{Stereo Camera:} StereoLabs ZED stereo vision camera providing real-time RGB and depth sensing, operating at 15~Hz, with a maximum field of view of 90° (H) × 60° (V) × 100° (D) and depth range of 0.5 m to 25 m. 

\item \textbf{GNSS:} Yoctopuce Yocto-GPS-V2 module based on the u-blox NEO-M8Q chipset with 72-channel multi-constellation support. It provides positioning updates at 10 Hz with a resolution of $10^{-6}$ degrees ($\sim$10 cm).

\item \textbf{IMU:} Yoctopuce Yocto-3D-V2, integrating a 3D accelerometer (±16 g, 0.001 g resolution), 3D gyroscope (±2000°/s, 0.1°/s resolution), and 3D magnetometer (±13 gauss, 0.01 gauss resolution). It also offers 100 Hz gyroscopic attitude estimation and outputs orientation in quaternion or Tait-Bryan angles.

\item \textbf{Lux Meter:} Yoctopuce Yocto-Light-V4 ambient light sensor, providing 0.01 lux sensitivity and measuring up to 83,000 lux at 10 Hz.

\item \textbf{Onboard Computer:} An NVIDIA Jetson AGX Orin Developer Kit (64 GB) served as the onboard computer, chosen for its high computational performance and integrated CUDA cores for real-time sensor processing and DL inference.
\end{itemize}

\noindent These sensors were acquired to support monitoring, temporal alignment, and dataset enrichment. However, only EEG data were used for model training and evaluation. EEG signals were recorded during the experiments using a wireless 16-channel OpenBCI Cyton and Daisy system with a gel-free electrode cap~\cite{openbci}, configured according to the international 10–20 system \cite{iyama2025high}. EEG signals were acquired at 125 Hz, and electrodes were positioned at Fp1, Fp2, F3, F4, F7, F8, C3, C4, T3, T4, T5, T6, P3, P4, O1, and O2 ~\cite{openbci}. These covered frontal, central, parietal, occipital, and temporal regions of the scalp. 

All robot sensors, along with the EEG headset and Xbox controller, interfaced as Robot Operating System (ROS-Noetic) nodes operating as followers, with the Jetson AGX Orin acting as the leader to manage time-stamped data publishing, subscription, and synchronisation within a ROS environment. Each modality published time-stamped messages on dedicated ROS topics aligned to a unified ROS time base, ensuring sub-millisecond synchronisation. No external synchronisation hardware was required. This architecture enabled real-time data acquisition, wireless communication, and autonomous logging during experiments.

\subsection{Data Labelling}
\label{sec:method:labelling}
The joystick's input from the Xbox controller was recorded by publishing messages to the relevant ROS topic at a rate of 10 Hz. Each message contained linear and angular velocity components, which were processed using a rule-based thresholding strategy to assign a discrete directional label (forward, reverse, left, right, stop), reflecting the participant’s intended movement. A small threshold $\tau$ was introduced to suppress noise around zero and ensure robust command detection. Let $v_x$ denote the linear velocity (forward/backward) and $\omega_z$ the angular velocity (turning):   

\[
\text{label} =
\begin{cases}
\text{forward}, & v_x > \tau \ \land \ |\omega_z| \leq \tau \\
\text{reverse}, & v_x < -\tau \ \land \ |\omega_z| \leq \tau \\
\text{left}, & \omega_z > \tau \ \land \ |v_x| \leq \tau \\
\text{right}, & \omega_z < -\tau \ \land \ |v_x| \leq \tau \\
\text{stop}, & |v_x| \leq \tau \ \land \ |\omega_z| \leq \tau \\
\end{cases}
\]
\noindent Samples with $|v_x| > \tau \ \land \ |\omega_z| > \tau$ were discarded due to inconsistency. Subsequently, each EEG sample at time $t$ was assigned the joystick label corresponding to the timestamp $t + \Delta$, where $\Delta \in \{0, 300, 400, 500, 600, 700, 800, 900, 1000\}$ ms. This formulation supports both real-time ($\Delta = 0$) and anticipatory decoding tasks for a predictive time horizon of $\Delta$. The labelling strategy is defined as $\text{Label}(t) = \text{Joystick}(t + \Delta)$.
This method generated one label for immediate action ($\Delta = 0$) and eight additional labels for future prediction ($\Delta > 0$). All labels were assigned using nearest-neighbour matching between EEG and joystick timestamps, preserving alignment accuracy within ROS time constraints.

\subsection{EEG Data Preparation Pipeline}
\label{sec:method:preparation}
EEG data is inherently non-stationary and varies over time, especially in real-world mobile navigation tasks. Key challenges include EEG drift (context- or fatigue-related signal changes)~\cite{zeng2022classifying} and BCI illiteracy (when some users cannot generate clear, classifiable neural patterns) \cite{lee2019eeg}. These effects may introduce class imbalance and temporal inconsistencies. To address them, we designed a temporal-aware, label-stratified data preparation pipeline that preserves class balance and chronological order during training and evaluation, reducing temporal leakage and ensuring that evaluation metrics better reflect real-world conditions:

\begin{algorithm}[t]
\setstretch{1.5} 
\small
\caption{\small Temporal-stratified EEG splitting and windowing.}
\label{alg:temporal}

\makebox[\linewidth][l]{\textbf{Input:} Preprocessed EEG data $X(t) \in \mathbb{R}^{C \times T}$, labels $y(t)$, }
\makebox[\linewidth][l]{\hspace*{3em}window size $S=125$, overlap $50\%$}
\textbf{where } $C$: number of EEG channels, $T$: number of time samples.
\makebox[\linewidth][l]{\textbf{Output:} Train/Test datasets $\mathcal{D}^{\text{train}}, \mathcal{D}^{\text{test}}$.}

\small
\begin{algorithmic}[1]
\STATE $\mathcal{D}_\ell = \{ (x_t,y_t) \mid y_t=\ell \}, \quad \forall \ell \in \mathcal{L}$ \hfill $\triangleright$ group by label
\STATE $\mathcal{D}_\ell = \sort_t(\mathcal{D}_\ell)$ \hfill $\triangleright$ chronological order
\STATE $\mathcal{D}_\ell = \bigcup_{i=1}^{N} \mathcal{C}_{\ell,i}, \quad N=100$ \hfill $\triangleright$ temporal chunks
\STATE $\mathcal{C}_{\ell,i}^{\text{train}} = \text{first } 0.7|\mathcal{C}_{\ell,i}|$ 
\STATE $\mathcal{C}_{\ell,i}^{\text{test}} = \text{last } 0.3|\mathcal{C}_{\ell,i}|$

\STATE $\begin{aligned}
\mathcal{D}^{\text{train}} &= \sort_t\Big(
   \bigcup_{\ell,i} \mathcal{C}_{\ell,i}^{\text{train}}
\Big)
\end{aligned}$

\STATE $\begin{aligned}
\mathcal{D}^{\text{test}} &= \sort_t\Big(
   \bigcup_{\ell,i} \mathcal{C}_{\ell,i}^{\text{test}}
\Big)
\end{aligned}$

\STATE $W_j = X[t_j : t_j+S-1], \quad t_{j+1} = t_j + S/2$ \\
\hfill $\triangleright$ sliding windows
\STATE $y_j = \arg\max_{\ell \in \mathcal{L}} \;\countop\{ y_t=\ell,\ t \in [t_j,t_j+S-1]\}$    \\
\hfill $\triangleright$  majority label
\STATE $\mathcal{D}^{\text{train}}_{\text{balanced}} = \OVRS(\mathcal{D}^{\text{train}})$ \hfill $\triangleright$ oversampling

\end{algorithmic}
\end{algorithm}

\begin{table*}[ht]
\centering
\caption{Deep learning models evaluated for EEG classification, grouped by architecture family.}
\label{tab:model_summary}
\scriptsize
\begin{tabular*}{\textwidth}{@{\extracolsep{\fill}}lllll@{}}
\hline
\textbf{Family} & \textbf{Model} & \textbf{Purpose} & \textbf{Focus} & \textbf{Strengths} \\
\hline
\multirow{7}{*}{CNN}
& EEGNet~\cite{lawhern2018eegnet}
& Compact EEG decoding
& Spatial-temporal features
& Robust to noise; lightweight \\
& DeepConvNet~\cite{schirrmeister2017deep}
& Deep feature extraction
& Spatial-temporal features
& Strong representation learning \\
& ShallowConvNet~\cite{schirrmeister2017deep}
& Short-window EEG decoding
& Temporal patterns
& Interpretable filters; effective for MI \\
& STNet~\cite{zhang2022ganser}
& Multi-channel EEG modelling
& Spatial-temporal dependencies
& Captures electrode interactions \\
& TSCeption~\cite{ding2022tsception}
& Frequency-aware EEG decoding
& Temporal-spectral features
& Multi-scale kernels \\
& CCNN~\cite{yang2018continuous}
& Real-time EEG classification
& Low-latency processing
& Lightweight; portable BCI suitability \\
& CNN1D~\cite{taghizadeh2024eeg}
& Time-series EEG decoding
& Temporal structure
& Efficient temporal extraction \\
\hline
\multirow{2}{*}{RNN}
& LSTM~\cite{zhang2021deep}
& Sequential EEG modelling
& Long-term dependencies
& Strong temporal memory \\
& GRU~\cite{zhang2021deep}
& Efficient temporal modelling
& Short/mid-term dependencies
& Fewer parameters \\
\hline
\multirow{2}{*}{Transformer}
& ViT~\cite{dosovitskiy2020image}
& Global EEG representation
& Long-range dependencies
& Attention-based global context \\
& EEGConformer~\cite{song2022eeg}
& Hierarchical EEG decoding
& Local-global features
& Combines CNN locality with attention \\
\hline
\end{tabular*}
\end{table*}

\subsubsection{Preprocessing}
EEG preprocessing was performed using the PyPREP pipeline \cite{bigdely2015prep} (MNE-Python implementation \cite{gramfort2013meg}) to provide robust and reproducible cleaning. The following steps were applied:
\begin{itemize}
    \item \textbf{High-pass filtering at 1 Hz:} Eliminates slow drifts and low-frequency trends caused by sweating, electrode shifts or baseline fluctuations to provide a stable signal for subsequent analysis.
    \item \textbf{Adaptive notch filtering at 50 Hz:} Suppresses powerline interference while preserving neural oscillations.
    \item \textbf{Robust average referencing:} Iteratively computed while excluding noisy channels, to reduce reference bias and improve spatial resolution.
    \item \textbf{Noisy channel detection and interpolation:} Channels with abnormal variance, low correlation, or RANSAC-detected artefacts were replaced using interpolation.
    \item \textbf{Z-score normalisation:} Each channel was normalised to zero mean and unit variance, minimising bias introduced by amplitude differences.
\end{itemize}
\vspace{0.5em}
This pipeline reduces non-stationary noise, line interference, and channel-level artefacts, thereby reducing the influence of physiological and motion-related contamination.
\vspace{0.5em}
\subsubsection{Temporal-stratified splitting}
After preprocessing, we implemented a temporal-aware label-stratified splitting procedure to prepare data for model training and evaluation. The procedure is outlined in Algorithm~\ref{alg:temporal}. Samples were grouped by label, sorted chronologically, and partitioned into 100 temporally contiguous chunks per label class. Within each chunk, a 70/30 split was applied to form training and testing sets. All chunk-wise partitions were aggregated and sorted globally by timestamp to maintain chronological integrity. After aggregation, samples were reordered according to their timestamps. While the train/test split introduces natural gaps in the sequence, the temporal order within each partition was preserved, avoiding artificial discontinuities and minimising distortion of EEG temporal dependencies. A sliding window of 125 samples (1 second) with 50\% overlap was applied, and each window was assigned the majority label of the samples it contained. Sliding windows were generated independently within the training and testing partitions to preserve temporal continuity, prevent leakage and minimise effects on EEG temporal and spectral characteristics. Random oversampling was applied only to the training set to mitigate class imbalance, while the testing set remained untouched to ensure unbiased evaluation. This strategy minimised temporal leakage while maintaining a realistic evaluation protocol.

\subsection{Model Training and Benchmarking}
\label{sec:method:model_train}
A diverse set of DL architectures, tailored for EEG decoding, was benchmarked to evaluate end-to-end classification performance without relying on manual feature extraction. The models spanned convolutional, recurrent, and transformer-based families, each selected for their demonstrated strengths in capturing spatial, temporal, and spectral characteristics of EEG signals. Table~\ref{tab:model_summary} summarises the core purpose and advantages of each model.
Deep convolutional networks such as EEGNet and STNet are known for their robustness to EEG noise and ability to generalise across subjects. Sequential models, including LSTM and GRU, are well-suited for capturing temporal dependencies in EEG time series, while architectures like ShallowConvNet and CCNN offer a favourable balance between classification performance and interpretability, which makes them applicable to medical and real-time systems. Lightweight models such as EEGNet and CCNN are particularly optimised for embedded applications, whereas transformer-based models like ViT and EEGConformer leverage attention mechanisms to enhance feature representation and classification accuracy.

Model training was conducted following a within-subject within-session approach, i.e., each model was trained and tested on data from a single participant, as well as from a single session, to ensure temporal consistency and avoid cross-session leakage. Training and testing were repeated across all participants and sessions in the dataset, as well as for all label types ($\Delta$). Performance results for each label type were aggregated across all participants and sessions for reporting.

All models were trained using standardised hyperparameters to ensure fair comparison. The batch size was 128, with 2,000 training epochs and a weighted cross-entropy loss function to address class imbalance. The Adam optimiser with a fixed learning rate of 0.001 was used. Training and inference were conducted on an NVIDIA RTX A6000 GPU to accelerate high-capacity training workloads.

\subsection{Evaluation Metrics}
\label{sec:method:metrics}
Model performance was evaluated using accuracy, precision, recall, and F1-score. Accuracy quantified overall correctness, while precision and recall measured class-specific predictive quality. The F1-score provided a balanced summary of precision and recall, which is particularly useful under class imbalance scenarios and was thus used as our main benchmark metric, reflecting both correctness and confidence of probabilistic predictions. All metrics were computed per subject and per session and were then averaged, with standard deviation reported to capture model stability. Performance was evaluated independently for each prediction horizon ($\Delta \in [0,1000]$ ms) to assess temporal robustness.

\begin{figure*}[t]
    \vspace{-2mm}
    \centering
    \begin{subfigure}[t]{0.30\textwidth}
        \centering
        \includegraphics[width=\linewidth]{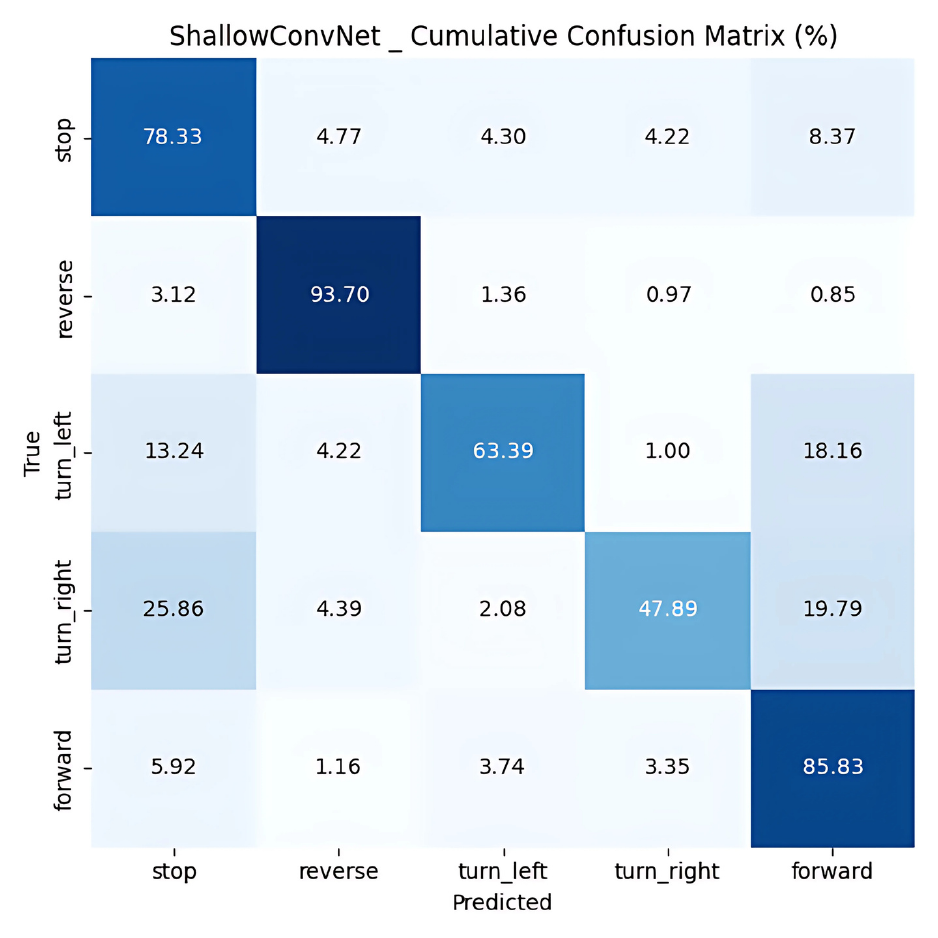}
        \subcaption*{(a)}
    \end{subfigure}
    \begin{subfigure}[t]{0.30\textwidth}
        \centering
        \includegraphics[width=\linewidth]{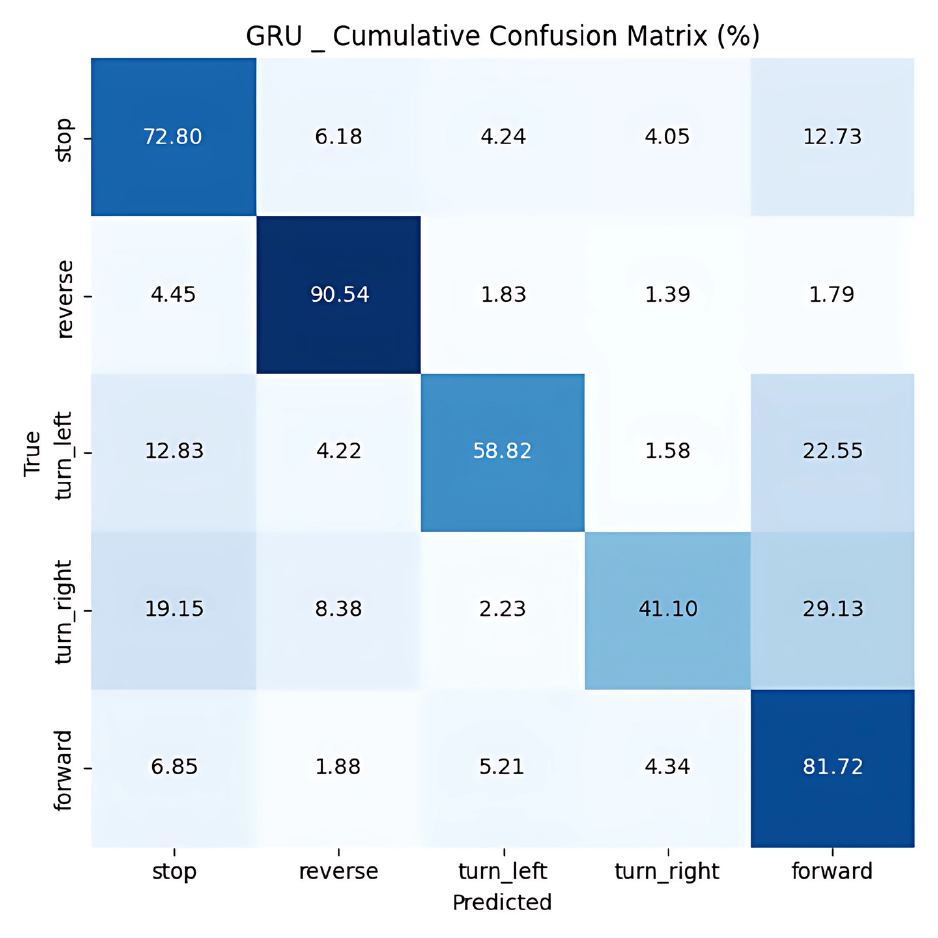}
        \subcaption*{(b)}
    \end{subfigure}
    \begin{subfigure}[t]{0.30\textwidth}
        \centering
        \includegraphics[width=\linewidth]{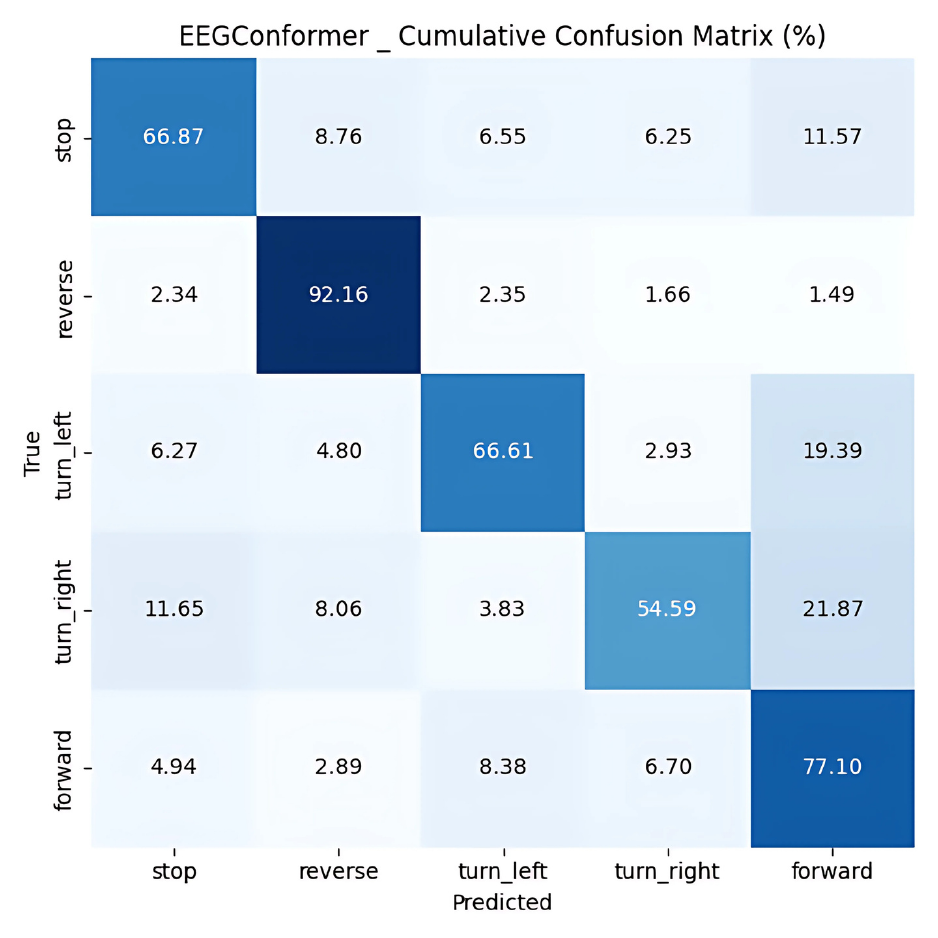}
        \subcaption*{(c)}
    \end{subfigure}

    \caption{Aggregated confusion matrices for (a) ShallowConvNet, (b) GRU, and (c) EEGConformer, 
    at $\Delta = 0$ ms.}
    \label{fig:confmat_label}
\end{figure*}

\begin{figure*}[t]
    \vspace{-2mm}
    \centering
    \begin{subfigure}[t]{0.30\textwidth}
        \centering
        \includegraphics[width=\linewidth]{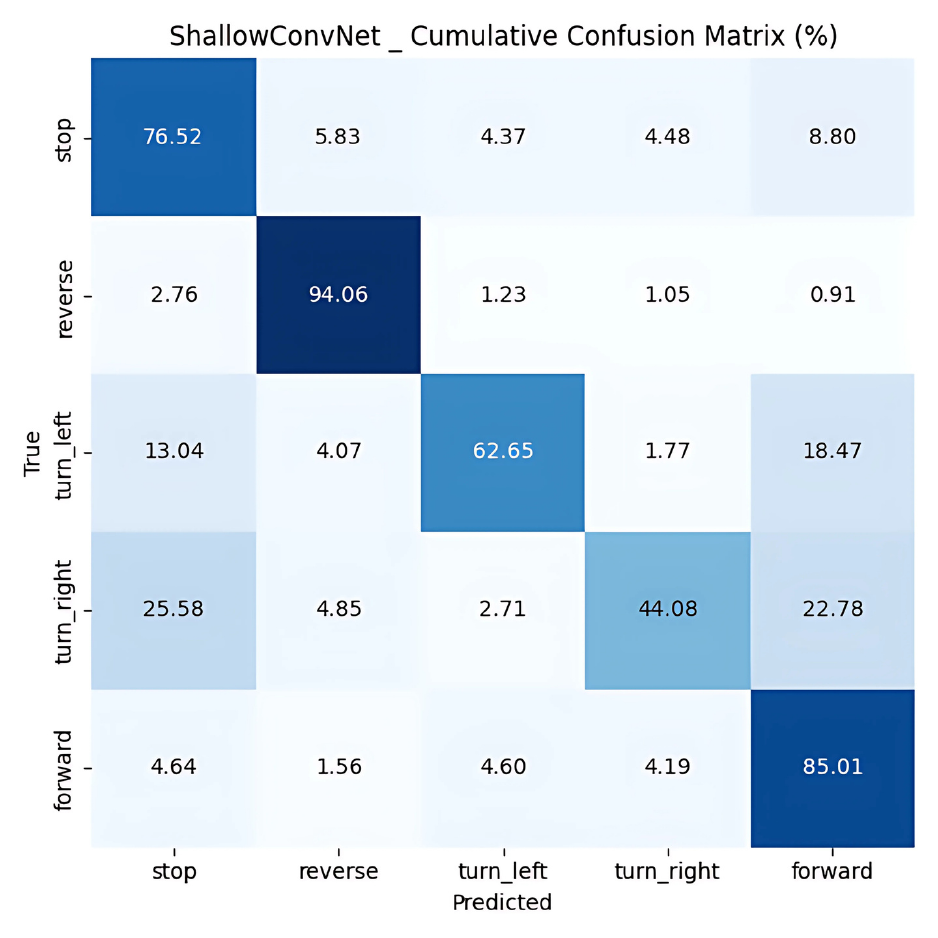}
        \subcaption*{(a)}
    \end{subfigure}
    \begin{subfigure}[t]{0.30\textwidth}
        \centering
        \includegraphics[width=\linewidth]{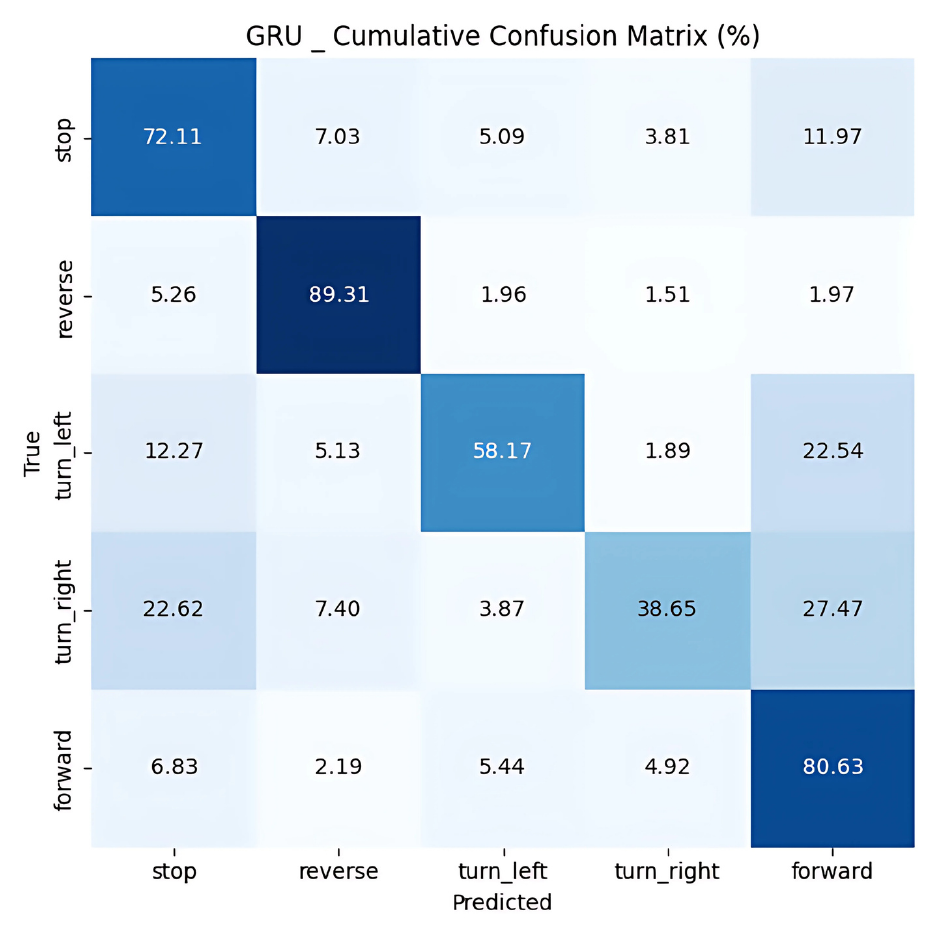}
        \subcaption*{(b)}
    \end{subfigure}
    \begin{subfigure}[t]{0.30\textwidth}
        \centering
        \includegraphics[width=\linewidth]{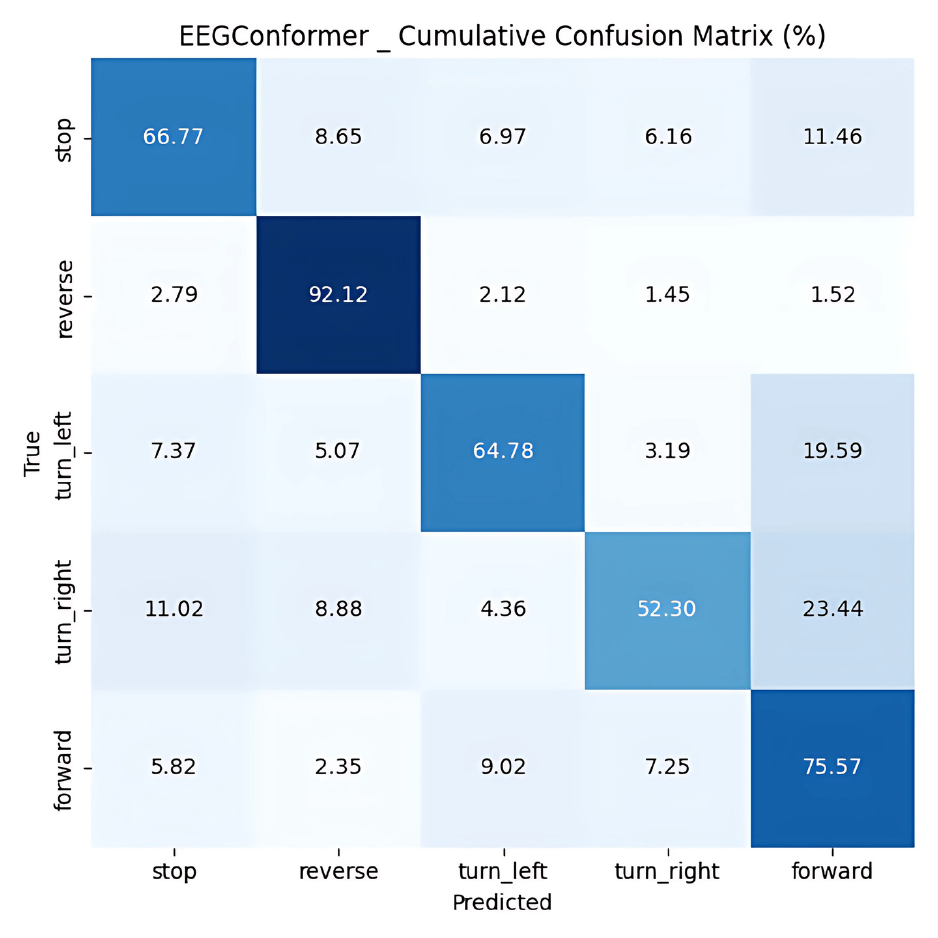}
        \subcaption*{(c)}
    \end{subfigure}

    \caption{Aggregated confusion matrices for (a) ShallowConvNet, (b) GRU, and (c) EEGConformer, 
    at $\Delta = 300$ ms.}
    \label{fig:confmat_300_three}
\end{figure*}

\begin{table*}[t]
\centering
\caption{Model performance comparison at $\Delta = 0$ ms (action) and $\Delta = 300$ ms (intention).}
\resizebox{\textwidth}{!}{%
\begin{tabular}{|l|l|cccc|cccc|}
\hline
& & \multicolumn{4}{c|}{\textbf{Action ($\mathbf{\Delta = 0}$ ms)}} & \multicolumn{4}{c|}{\textbf{Intention ($\mathbf{\Delta = 300}$ ms})} \\
\textbf{Family} & \textbf{Model} & \textbf{Accuracy} & \textbf{Precision} & \textbf{Recall} & \textbf{F1} & \textbf{Accuracy} & \textbf{Precision} & \textbf{Recall} & \textbf{F1} \\
\hline
\multirow{7}{*}{CNN} 
 & EEGNet         & 0.78 $\pm$ 0.13 & 0.63 $\pm$ 0.11 & 0.70 $\pm$ 0.13 & 0.64 $\pm$ 0.12 & 0.77 $\pm$ 0.13 & 0.62 $\pm$ 0.10 & 0.68 $\pm$ 0.13 & 0.62 $\pm$ 0.12 \\
 & DeepConvNet    & 0.33 $\pm$ 0.17 & 0.35 $\pm$ 0.14 & 0.35 $\pm$ 0.10 & 0.24 $\pm$ 0.12 & 0.32 $\pm$ 0.16 & 0.35 $\pm$ 0.14 & 0.35 $\pm$ 0.10 & 0.24 $\pm$ 0.12 \\
 & ShallowConvNet & \textbf{0.83 $\pm$ 0.11} & \textbf{0.67 $\pm$ 0.10} & \textbf{0.72 $\pm$ 0.12} & \textbf{0.67 $\pm$ 0.11} & \textbf{0.82 $\pm$ 0.11} & \textbf{0.65 $\pm$ 0.09} & \textbf{0.71 $\pm$ 0.12} & \textbf{0.66 $\pm$ 0.11} \\
 & STNet          & 0.61 $\pm$ 0.15 & 0.46 $\pm$ 0.13 & 0.47 $\pm$ 0.14 & 0.46 $\pm$ 0.13 & 0.61 $\pm$ 0.15 & 0.46 $\pm$ 0.13 & 0.47 $\pm$ 0.14 & 0.45 $\pm$ 0.13 \\
 & TSCeption      & 0.81 $\pm$ 0.11 & 0.62 $\pm$ 0.11 & 0.63 $\pm$ 0.12 & 0.62 $\pm$ 0.11 & 0.81 $\pm$ 0.10 & 0.63 $\pm$ 0.11 & 0.62 $\pm$ 0.11 & 0.62 $\pm$ 0.11 \\
 & CCNN           & 0.61 $\pm$ 0.33 & 0.45 $\pm$ 0.30 & 0.53 $\pm$ 0.27 & 0.46 $\pm$ 0.30 & 0.60 $\pm$ 0.32 & 0.44 $\pm$ 0.29 & 0.52 $\pm$ 0.26 & 0.45 $\pm$ 0.29 \\
 & CNN1D          & 0.79 $\pm$ 0.11 & 0.61 $\pm$ 0.11 & 0.62 $\pm$ 0.12 & 0.61 $\pm$ 0.11 & 0.79 $\pm$ 0.11 & 0.60 $\pm$ 0.11 & 0.61 $\pm$ 0.12 & 0.60 $\pm$ 0.11 \\
\hline
\multirow{2}{*}{RNN}
 & GRU            & \textbf{0.78 $\pm$ 0.13} & \textbf{0.63 $\pm$ 0.11} & \textbf{0.67 $\pm$ 0.13} & \textbf{0.63 $\pm$ 0.13} & \textbf{0.77 $\pm$ 0.15} & \textbf{0.62 $\pm$ 0.12} & \textbf{0.66 $\pm$ 0.14} &\textbf{0.62 $\pm$ 0.13} \\
 & LSTM           & 0.71 $\pm$ 0.22 & 0.57 $\pm$ 0.19 & 0.62 $\pm$ 0.19 & 0.57 $\pm$ 0.20 & 0.71 $\pm$ 0.21 & 0.57 $\pm$ 0.18 & 0.62 $\pm$ 0.18 & 0.57 $\pm$ 0.19 \\
\hline
\multirow{2}{*}{Transformer}
 & ViT            & 0.70 $\pm$ 0.12 & 0.53 $\pm$ 0.11 & 0.53 $\pm$ 0.11 & 0.52 $\pm$ 0.11 & 0.69 $\pm$ 0.12 & 0.52 $\pm$ 0.10 & 0.53 $\pm$ 0.11 & 0.52 $\pm$ 0.10 \\
 & EEGConformer   & \textbf{0.75 $\pm$ 0.13 }& \textbf{0.61 $\pm$ 0.10} & \textbf{0.69 $\pm$ 0.13} & \textbf{0.61 $\pm$ 0.12} &\textbf{0.74 $\pm$ 0.13} & \textbf{0.60 $\pm$ 0.10} & \textbf{0.68 $\pm$ 0.13} & \textbf{0.60 $\pm$ 0.11} \\
\hline
\multicolumn{10}{l}{\scriptsize \textsuperscript{*}Results in bold indicate best performance per model family.} \\
\end{tabular}%
}
\label{tab:model_performance}
\end{table*}

\section{Results and Discussion}
The average performance of the 11 examined DL models across all subjects and sessions was evaluated across nine labelling conditions corresponding to the executed action ($\Delta = 0$ ms) and predictive horizons up to 1000 ms, as explained in Section~\ref{sec:method:labelling}. Table~\ref{tab:model_performance} shows detailed results for all models at $\Delta = 0$ ms and $\Delta = 300$ ms, whereas \figurename~\ref{fig:f1} illustrates the performance of all examined models across all labelling horizons. ($\Delta = 300$ ms) was selected as a representative anticipatory horizon, reflecting a balance between early intention prediction and classification stability.

\subsection{Main findings}

\subsubsection{CNN-based architectures emerged as the most robust and accurate models}
At $\Delta = 0$  ms, ShallowConvNet achieved the best results, with a mean accuracy of 83\% and F1-score of 67\%, outperforming both recurrent and transformer-based approaches. EEGNet (accuracy = 78\% / F1 = 64\%) and CNN1D (accuracy = 79\% / F1 = 61\%) also achieved competitive results, further validating the strength of compact convolutional filters for extracting discriminative spatio-temporal patterns in EEG. Importantly, ShallowConvNet remained the top performer across the examined predictive horizons, achieving an F1-score of 66\% at  $\Delta = 300$ ms, followed closely by EEGNet (62\%), TSCeption (62\%) and GRU (62\%). The confusion matrices for ShallowConvNet at $\Delta = 0$  ms (\figurename~\ref{fig:confmat_label}a) and $\Delta = 300$ ms (\figurename~\ref{fig:confmat_300_three}a) clearly illustrate its stability across classes.

\subsubsection{Other examined CNNs highlighted trade-offs}
DeepConvNet showed the lowest performance (accuracy = 32\% /  F1 = 24\%  at $\Delta = 300$ ms), suggesting over-parameterisation and overfitting in this application. Meanwhile, temporal–spectral hybrids such as TSCeption (accuracy = 81\% /  F1 = 62\%) and STNet (accuracy = 61\% /  F1 = 45\%) benefited from multi-scale filtering, achieving competitive mid-range results but not surpassing the CNN baselines.

\subsubsection{Recurrent models demonstrated complementary strengths}
GRU, in particular, achieved results close to the best CNN at shorter horizons (accuracy = 78\% /  F1 = 63\% at  $\Delta = 0$ ms; accuracy = 77\% /  F1 = 62\% at  $\Delta = 300$ ms), suggesting that its gating mechanisms effectively capture short-term temporal dependencies in EEG. However, recurrent models exhibited greater variability across subjects and horizons compared to CNNs, likely due to their sensitivity to temporal noise and subject-specific EEG dynamics. The GRU's confusion matrices for $\Delta = 0$ ms and for $\Delta = 300$ ms appear in \figurename~\ref{fig:confmat_label}b and \figurename~\ref{fig:confmat_300_three}b, respectively.

\pgfplotstableread[row sep=\\,col sep=&]{
Delta &	CCNN &	CNN1D &	DeepConvNet &	EEGConformer &	EEGNet &	GRU &	LSTM &	ShallowConvNet &	STNet &	TSCeption	& ViT \\
0	& 0.46	& 0.61	&0.24	&0.61	&0.64	&0.63	&0.57	&0.67	&0.46	&0.62	&0.52 \\
300	& 0.45	& 0.6	&0.24	&0.6	&0.62	&0.62	&0.57	&0.66	&0.45	&0.62	&0.52 \\
400	& 0.47	&0.6	&0.24	&0.6	&0.62	&0.62	&0.57	&0.66	&0.45	&0.61	&0.52 \\
500	& 0.47	&0.6	&0.24	&0.6	&0.62	&0.62	&0.56	&0.65	&0.44	&0.6	&0.51 \\
600	& 0.46	&0.59	&0.23	&0.59	&0.62	&0.61	&0.57	&0.65	&0.45	&0.61	&0.52 \\
700	& 0.46	&0.59	&0.24	&0.59	&0.62	&0.61	&0.56	&0.65	&0.44	&0.6	&0.51 \\
800	& 0.45	&0.59	&0.24	&0.58	&0.61	&0.61	&0.56	&0.64	&0.44	&0.6	&0.51 \\
900	& 0.43	&0.59	&0.23	&0.58	&0.61	&0.61	&0.56	&0.64	&0.44	&0.6	&0.51 \\
1000 &	0.45	&0.58	&0.24 &0.58	&0.6	&0.6	&0.56	&0.64	&0.44	&0.6	&0.51 \\
}\resultsdata

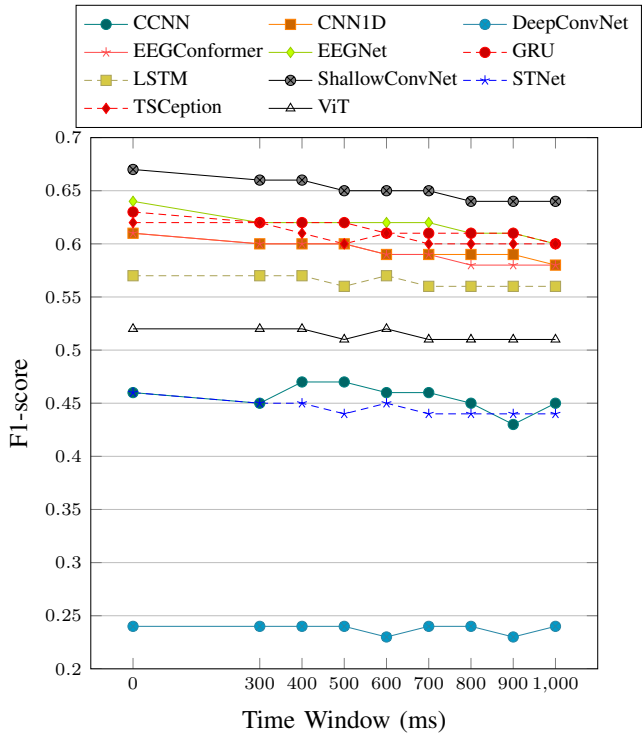
\begin{figure}[!t]
\centering
\resizebox{\columnwidth}{!}{%
\begin{tikzpicture}
\begin{axis}[
    width=0.99\columnwidth,
    height=0.50\textwidth,
    ymin=0.2,
    ymax=0.7,
    ytick distance=0.05,
    ymajorgrids,
    ylabel={F1-score},
    xlabel={Time Window (ms)},
    legend columns=3,
    legend cell align={left},
    xtick=data,
    legend style = {font = {\fontsize{8 pt}{12 pt}\selectfont},at={(-0.03,1.25)},anchor=north west},
    cycle list name=exotic, 
    tick label style={font=\scriptsize}
]
    \addplot table[x=Delta,y=CCNN] \resultsdata;
    \addplot table[x=Delta,y=CNN1D] \resultsdata;
    \addplot table[x=Delta,y=DeepConvNet] \resultsdata;
    \addplot table[x=Delta,y=EEGConformer] \resultsdata;
    \addplot table[x=Delta,y=EEGNet] \resultsdata;
    \addplot table[x=Delta,y=GRU] \resultsdata;
    \addplot table[x=Delta,y=LSTM] \resultsdata;
    \addplot table[x=Delta,y=ShallowConvNet] \resultsdata;
    \addplot table[x=Delta,y=STNet] \resultsdata;
    \addplot table[x=Delta,y=TSCeption] \resultsdata;
    \addplot[mark=triangle] table[x=Delta,y=ViT] \resultsdata;
    \legend{CCNN,CNN1D,DeepConvNet,	EEGConformer, EEGNet, GRU, LSTM, ShallowConvNet, STNet, TSCeption, ViT};
\end{axis}
\end{tikzpicture}
} 
\caption{F1-scores across all labelling horizons ($\Delta \in [0,1000]$ ms) for all examined models.}
\label{fig:f1}
\end{figure}

\subsubsection{Transformer-based approaches revealed mixed outcomes}
EEGConformer maintained stable performance across horizons (accuracy = 75\% / F1 = 61\% at $\Delta = 0$; accuracy = 74\% / F1 = 60\% at $\Delta = 300$ ms), reflecting the benefit of combining convolution for local feature extraction with attention for refinement. In contrast, ViT's performance dropped more noticeably at $\Delta = 300$ ms (accuracy = 69\% / F1 = 52\%), as it relies only on attention, and attention models typically require large amounts of data to train well.
\figurename~\ref{fig:confmat_label}c and \figurename~\ref{fig:confmat_300_three}c illustrate EEGConformer's confusion matrices for $\Delta = 0$ ms and for $\Delta = 300$ ms, respectively.

\subsubsection{Across all models, the F1-vs-prediction-horizon curves (\figurename~\ref{fig:f1}) confirmed a gradual decline as the prediction horizon increased} 
Despite this decrease, ShallowConvNet, EEGNet, and GRU consistently retained F1-scores above 60\% up to $\Delta = 900$ ms, indicating their robustness for anticipatory decoding. This plateau suggests that brain–robot interfaces can reliably anticipate user actions within a 0.9-second window, a timescale well aligned with real-time navigation and control requirements.

\subsection{Further Discussion}
Table~\ref{tab:model_performance} summarises performance at $\Delta = 0$ ms and $\Delta = 300$ ms across all 11 models and concisely captures the aforementioned trends.  In particular, CNN-based models dominate overall accuracy, while the GRU demonstrates relatively stable performance across prediction horizons (\figurename~\ref{fig:f1}), likely due to its recurrent memory. The transformer baseline remains consistent yet moderate, as attention models are data-hungry and better suited to larger datasets. Overall, the results demonstrate that compact CNNs provide the best trade-off between performance, robustness and efficiency, outperforming recurrent and transformer models in anticipatory EEG decoding. The GRU remains a recurrent benchmark, while EEGConformer highlights the potential of transformer-based hybrids with larger datasets. Underperforming models, such as DeepConvNet and ViT, highlight the risks of over-complexity and data hunger in limited-sample EEG contexts. Collectively, these findings establish baselines and confirm the viability of convolutional EEG decoders for real-time brain–robot control.

\section{CONCLUSIONS}
This study presented a real-world benchmark for EEG-driven intention decoding in brain–robot interfaces, combining multi-command control on a robotic rover with evaluation of 11 DL models. By introducing action-aligned ($\Delta = 0$ ms) and predictive labelling horizons ($\Delta \in [0,1000]$ ms), we showed that compact CNNs, particularly ShallowConvNet, achieved the best performance, while GRU provided a strong recurrent baseline. The transformer-based EEGConformer offered consistent performance, whereas ViT and DeepConvNet underperformed due to data demands or over-parameterisation. Several models retained F1-scores above 60\% up to $\Delta = 900$ ms, which confirms anticipatory EEG decoding's feasibility. These findings were derived from multi-session experiments spanning 120 sessions, with results reflecting session variability and robustness.

Future work will extend this framework to online experiments, cross-session robustness investigation, cross-subject generalisation, and multimodal fusion. Pre-training and data-efficient transformer variants may further enhance attention-based models.  Overall, the proposed benchmark establishes reproducible baselines, clarifies architectural trade-offs, and provides groundwork for predictive BCI/BCV navigation.



\bibliographystyle{bib/IEEEtran.bst} 
\bibliography{bib/main}

\end{document}